\documentclass[conference]{IEEEtran}
\IEEEoverridecommandlockouts
\usepackage{cite}
\usepackage{amsmath,amssymb,amsfonts}
\usepackage{algorithmic}
\usepackage{graphicx}
\usepackage{textcomp}
\usepackage{xcolor}
\def\BibTeX{{\rm B\kern-.05em{\sc i\kern-.025em b}\kern-.08em
    T\kern-.1667em\lower.7ex\hbox{E}\kern-.125emX}}

\usepackage{fancyhdr}
\begin{document}

\title{Modeling Player Personality Factors from In-Game Behavior and Affective Expression\\
\thanks{This work is funded by National Science Foundation (NSF), Division of Information and Intelligence Systems (IIS) under Grant Number 1355298 awarded to the project V-PAL: Virtual Personality Assessment
Laboratory.}
}

\author{\IEEEauthorblockN{Reza Habibi}
\IEEEauthorblockA{\textit{Computational Media Department} \\
\textit{University of California, Santa Cruz}\\
Santa Cruz, USA\\
rehabibi@ucsc.edu}
\and
\IEEEauthorblockN{Johannes Pfau}
\IEEEauthorblockA{\textit{Computational Media Department} \\
\textit{University of California, Santa Cruz}\\
Santa Cruz, USA\\
jopfau@ucsc.edu}
\and
\IEEEauthorblockN{Magy Seif El-Nasr}
\IEEEauthorblockA{\textit{Computational Media Department} \\
\textit{University of California, Santa Cruz}\\
Santa Cruz, USA\\
mseifeln@ucsc.edu}
}

\maketitle
\thispagestyle{fancy}

\begin{abstract}
Developing a thorough understanding of the target audience (and/or single individuals) is a key factor for success - which is exceptionally important and powerful for the domain of video games that can not only benefit from informed decision making during development, but ideally even tailor game content, difficulty and player experience while playing. The granular assessment of individual personality and differences across players is a particularly difficult endeavor, given the highly variant human nature, disagreement in psychological background models and because of the effortful data collection that most often builds upon long, time-consuming and deterrent questionnaires. In this work, we explore possibilities to predict a series of player personality questionnaire metrics from recorded in-game behavior and extend related work by explicitly adding affective dialog decisions to the game environment which could elevate the model's accuracy. Using random forest regression, we predicted a wide variety of personality metrics from seven established questionnaires across ($n=62$) players over 60 minute gameplay of a customized version of the role-playing game \textit{Fallout: New Vegas}. While some personality variables could already be identified from reasonable underlying in-game actions and affective expressions, we did not find ways to predict others or encountered questionable correlations that could not be justified by theoretical background literature. Yet, building on the initial opportunities of this explorative study, we are striving to massively enlarge our data set to players from an ecologically valid industrial game environment and investigate the performance of more sophisticated machine learning approaches.
\end{abstract}


\section{Introduction}
Video games have become increasingly popular over the past few decades \cite{b0}, and this rise in popularity goes along with an opportunity to explore how players behave and interact with and in video game environments. This can not only aid development in being able to closely tailor game content or narrative to the target group or individual, but also opens possibilities to make avail of intrinsically motivating game settings and detailed means of behavioral assessment for the sake of investigating psychological patterns beyond gaming. One particularly attractive area of interest is the reasoning over players' in-game actions which can provide systematic insights into their personalities or affective behaviors. Vice versa, individual differences in personality have been found to be predictive of in-game actions, suggesting that there may be a reciprocal relationship between in-game actions and personality \cite{b9}. By understanding these relationships between in-game actions and personality, we can gain insights into the complexities of affective behaviors in players, provide more nuanced understandings of how individuals engage with and respond to video games, and can implement or elevate player-centeric adaptation technologies within gameplay or during development.

Recently, the use of video games in the field of affective computing has garnered broad attention. Games have been used to explore various aspects of affect, such as emotion \cite{b14,b17}, player experience \cite{b15}, empathy \cite{b16}, and other related areas \cite{b18,b19,b20,b21}. By integrating game elements and mechanics, such as immersive graphics, captivating storylines, and interactive gameplay, video games offer a conducive environment to investigate affective computing \cite{b22,b23,b24}. However, developing games to specifically study individual differences and affect-related behaviors comes with distinct challenges about measurement and applicability \cite{b1a}, which we strive to extend with this work.

Digital games are designed to engage users in various activities, such as simulations, problem-solving tasks, and decision-making exercises, making them valuable resources for understanding human behavior and training models \cite{b2,b3,b4,blascovich}. While there exist various versions of serious games, not all of them are designed to study human behavior or gather data. However, some versions possess the capability to establish controlled and realistic environments that enable users to interact with. This aspect constitutes a significant advantage of serious games. One of the advantages of serious games is their potential to emulate intricate scenarios that may be challenging or unfeasible to replicate in real-life situations \cite{b1b,b1c}. Ideally, they are also effective tools for collecting large amounts of data on user behavior and can attract more people to interact with them due to their ideally motivating and enjoyable nature \cite{b1d}. As users interact with the game, they generate data that can be analyzed to identify patterns and trends in human behavior. Furthermore, serious games are highly customizable, allowing researchers to manipulate various aspects and factors to study their impact on human behavior using different layers of information such as communication, game-specific metrics, or player activities.

Individual differences refer to the variations in personality, cognition, behavior, and other characteristics that exist between people \cite{b24}. These differences can manifest in various ways, such as learning style, problem-solving, emotional intelligence, and social skills. Individual differences can also influence how people express, regulate or experience emotions \cite{b25,b28,b29,b30}. Therefore, understanding individual differences is critical in studying affect, as it allows researchers in the field to identify factors that contribute to emotional variability and create tailored interventions to better meet the needs of individuals\cite{b26,b27}. In the five-factor model of personality (FFM) \cite{b28b}, one of the currently most established approaches, there are different traits (domains) related to different aspects of emotion and affect in general. For instance, the trait of neuroticism is strongly related to affective behaviors. Individuals who score high in neuroticism are more likely to experience negative emotions such as anxiety, sadness, and anger, and may have difficulty regulating their emotions. On the other hand, individuals who score low in neuroticism tend to be more emotionally stable and better at coping with stressors\cite{b28}. In contrast, extraversion depicts another personality trait related to the tendency to experience more positive emotions \cite{b30}.

This work hypothesizes that tracking player action choices over time in a role-playing game (RPG) can provide insights into their individual differences and disposition to affect-related behaviors. We focus on predicting players' individual personality differences
from their interest in pursuing higher-level RPG activities (such as Quests, Combat, NPC and Inventory actions) in an open-world setting, and additionally from their active expression of affective dialog choices in NPC conversations (e.g. supportive, aggressive or apprehensive dialog lines) \cite{cano}.

In addition to recorded in-game behavior, we collected broad perspectives on participants' personality through multiple established questionnaires, including the Revised NEO Personality Inventory (NEO PI-R) \cite{b28c}, the Ego-Resilience Scale (ER 89) \cite{b28f}, the Attributional Complexity Scale (ACS) \cite{b28d}, the Buss-Perry Aggression Questionnaire (BPAQ) \cite{b28e}, among others.
Regarding the logged game data, we created an aggregated dataset over all actions of individual players over time that represented their interest in pursuing higher-level actions of the open-world RPG. Additionally, we designed all conversations that players would hold with NPCs so that they offer distinct affective choices (based on Plutchik's Wheel of Emotions \cite{plutchik}, following related work on a taxonomy of affect for chat conversation \cite{b28j}).
After gameplay, we applied random forest regression to our data to process the player's actions and decisions and predicted underlying personality factors for both data types.

With preliminary results of this exploratory investigation \cite{cano}, we aim to answer the following research questions and open up the path for more advanced machine learning models and more comprehensive data sets: (a) Can we predict players' individual personality differences from recorded in-game behavior? (b) Which in-game actions or decisions constitute the most important features in those individual differences? In order to answer these questions we utilized machine learning methods on VPAL in-game behavior data \cite{cano,GDS}, as well as a broad range of established personality questionnaires, to triangulate and predict detailed facets of players' personalities.

\section{Related work}
Recently, there has been significant interest in the modeling and prediction of Individual differences and facets related to affect \cite{b32}. Prior research has utilized artificial intelligence and serious games to improve the understanding, modeling, and prediction of these differences and behaviors. In this section, we provide a brief overview of the modeling of individual differences and affect-related facets and highlight some of the challenges and limitations of existing methods. We then review the research that has employed AI and serious games to better understand, model, and predict affect related facets in individual differences. Specifically, we discuss the different approaches used in these studies and the insights gained from their findings.

Games are an area of growing interest in affective computing and other computational sciences. Simulations, problem-solving tasks, and decision-making exercises, as well as providing an environment to evoke emotional experiences, are a few examples of activities that deeply engage players, making them valuable resources for comprehending human behavior and training models accordingly \cite{b2,b3,b4}. (Serious) games possess the potential to customize every aspect of their design to tailor specific affective behaviors \cite{b5,b6}, and also a great environment to model players behaviors. Yee et al. analyzed the behavior of 1040 players in \textit{World of Warcraft}  \cite{b33}. They utilized in-game activities such as the number of kills and the role played, among others, and employed a linear regression approach to correlate actual and predicted FFM scores. However, they relied on the features they observed from players to be the most correlated with personality scores, which is not a good indicator of predicting personality. Another issue with this work is the lack of design and interaction to cover different personality types. Yaakub et al. \cite{b34} developed a simple 2D puzzle game and used the Myers Briggs Type Indicators (MBTI) to capture the personalities of 50 players. They employed the game's functionality to evaluate the user's personality, based on four types of personality categorized by MBTI as introverted or extroverted, and sensing, intuition, feeling, and thinking. The authors reported an accuracy rate of 77.5\% for capturing personality. However, the paper did not provide sufficient details on the evaluation and experiment procedures, as well as detailed results for personality facets.

Weber, and Mateas \cite{b8} utilized a series of classification algorithms, such as linear regression, to predict player strategies. Volkmar et al. tailored in-game achievements to individual differences and measured an increase in player experience if matching properly \cite{b8b}. Teng et al. used player journey map segmentation to investigate differences in gameplay based on -- or influencing -- higher-level metrics, which are not limited to personality variables \cite{b8c,b8d}. Habibi et al. measured differences in physiological responses between different personalities, especially the higher impact of stress on more extroverted persons \cite{b8e}. Kashani et al. investigated individual personality differences and their impact on in-game communication \cite{b8f}. Bunian et al. \cite{b9} employed Hidden Markov Modeling (HMM) to model individual differences using VPAL dataset. They generated behavioral features that were used to classify real-world players' characteristics, such as game expertise and the FFM personality traits. Their findings indicate that consciousness is a more predictive factor than others in their dataset. Chen et al. \cite{b35} employed VPAL data and sequential data to predict real-world characteristics such as gender, gameplay expertise, and five personality traits. They utilized sequential mining to extract close sequential patterns with over 40\% support in all action sequences in the game and used logistic regression to model the data. The authors reported accuracies of 52.4\% for Agreeableness, 47.8\% for Openness, 25\% for gameplay expertise, 26.1\% for Extraversion, 27.8\% for Neuroticism, 29.4\% for Conscientiousness, and 37.5\% for gender. However, their results may have been impacted by other factors, such as gameplay expertise and demographic data, which were not solely based on in-game activities.

The modeling and predicting affect and affective behaviors is a dynamic and active field within affective science, artificial intelligence, and game research. Picard \cite{b1} has stressed the significance of cultivating affective awareness in artificial intelligence has been underscored in diverse domains, including the realm of education. Adding to the advancements of existing literature, we utilized machine learning methods on in-game behavior data, as well as a broad range of established personality questionnaires, to triangulate and predict detailed facets of players' personalities.

\section{Design and Methodology}
\subsection{Game Environment}
To allow players to express their individual differences and affective behaviors by interacting with various possibilities, an immersive environment is needed that ideally resembles real-world scenarios. For this purpose, we utilized the Virtual Personality Assessment Lab (VPAL) \cite{cano,GDS}, which is a custom-modded playable subset of Fallout: New Vegas with open-world characteristics that facilitate unconstrained play. VPAL features short, custom scenario sets in an occupied western-style town that yields multiple adversarial encounters, quests and events. Players are open to respond to these situations in a variety of ways, such as exploring different areas like hotels and houses, collecting various objects and weapons, interacting with non-player characters (NPCs) using different choices, and participating in a range of quests and combat scenarios. Players embark on their journey through the game by visiting 
an introductory house with multiple rooms designed to help them familiarize themselves with the game's controls and become accustomed to the game mechanics. 
Beyond that, players are free to explore the open world, visit and search buildings, talk with NPCs, accept up to 15 different quests, collect, steal or craft items and can engage in combat (or be attacked) with any NPC.

\subsection{Behavioral Metrics}
To provide suitable metrics for predicting individual differences, the mechanics offered in the game were grouped into homogeneous categories of similar affordances. VPAL offers four different classes of metrics such as:
\begin{itemize}
    \item \textit{Combat Behavior}: This class captures aggressive interaction and attitude by allowing players to make unmotivated attacks on NPCs and creatures, as well as engage in combat initiated by NPCs against the player.
    \item \textit{Narrative}: This class captures eight behaviors which are aligned with completing tasks assigned to the player such as quests accepted or completed, quest-related dialog choices, and the total action steps they underwent for each quest.
    \item \textit{Conversation Interactions}: This class comprises six types of interactions with NPCs, including dialog choices for making small talk, time spent in dialogue, instances of dialogue, and time spent in the proximity of NPCs.
    \item \textit{Navigation and Interaction with the game world}: This class comprises nine behaviors, including (a) number of areas entered, (b) interactions with doors, (c) dialog choices inquiring about the world, (d) total distance traveled, (e) total head movement, (f) interactions with dead creatures and NPCs, (g) interactions with containers, and (h) objects activated by the player.
\end{itemize}

We utilized these categories to define different actions in game, which is presented in Table \ref{tab:in-game}.
\begin{table}[h]
\centering
\caption{Recorded in-game activity types (aggregated per player for prediction)}
\label{tab:in-game}
\begin{tabular}{|r|p{5.3 cm}|}
\hline
\textbf{Activity} & \textbf{Description} \\
\hline
Attack & The player successfully hit an NPC.\\
Attacked & The player got hit by an NPC attack.\\
AttackedFirst & The player initiated an attack on an NPC.\\
ObjectOnActivate & The player activated an object in the game, such as a switch or lever.\\
PlayerDropped & The player dropped an item from their inventory.\\
PlayerLootedItem & The player took an item from a lootable container the ground.\\
PlayerLootedDead & The player took items from a dead NPC's inventory.\\
PlayerEquipped & The player equipped an item from their inventory.\\
PlayerPickpocket & The player attempted to steal an item from an NPC.\\
PlayerShot & The player used a ranged weapon to attack an NPC.\\
PlayerKilled & The player killed an NPC.\\
InteractionNPC & The player interacted with an NPC in the game.\\
Quest & The player completed a quest or objective.\\
Dialogue & The player engaged in a conversation with an NPC.\\
PlayerJumped & The player made a jump in the game.\\
Attacked & The player was hit by an enemy NPC.\\
InteractionObject & The player interacted with an object in the game, such as a door or chest.\\
CraftingTable & The player used a crafting table to create an item.\\
PlayerDroppedItem & The player dropped an item from their inventory.\\
PlayerSold & The player sold an item to an NPC.\\
PlayerUsed & The player used an item from their inventory.\\
PlayerSneaking & The player moved around while crouched or sneaking.\\
PlayerShootinCorpse & The player shot a dead body.\\
\hline
\end{tabular}
\end{table}


\begin{table*}[!t]
\centering
\caption{Most accurate personality variable predictions after random forest regression. We report the explained variance R², the mean squared error and the three most important features among in-game actions that predicted a factor. For the latter, we indicate the direction of correlation with the target variable.}
\label{tab:results}
\begin{tabular}{|r|l|l|l|l|l|}
\hline
\multicolumn{1}{|c|}{\textbf{Scale}} &
\multicolumn{1}{|c|}{\textbf{Factor}} & \multicolumn{1}{c|}{\textbf{Description}} & \multicolumn{1}{c|}{\textbf{R²}} & \multicolumn{1}{c|}{\textbf{MSE}} &
\multicolumn{1}{c|}{\textbf{Highest Feature Importances}} 
\\
\hline
FFM & O & Openness & 0.33 & 214.80 & $\downarrow$ PlayerSneaking, $\downarrow$ AttackedFirst\\
RBQ & 36 & Expresses hostility & 0.30 & 0.40 & $\uparrow$ Quest, $\uparrow$ PlayerUsed \\
BPAQ & Verbal & Verbal Aggression & 0.28 & 14.72 & $\downarrow$ PlayerLootedItem, $\uparrow$ PlayerEquipped, $\uparrow$ PlayerDroppedItem\\
DSM V & RigidPerfection & Pursues perfection rigidly & 0.26 & 36.68 &  $\downarrow$ PlayerDroppedItem,  $\uparrow$ InteractionObject,  $\uparrow$ PlayerLootedItem \\
RBQ & 54 & Offers advice & 0.26 & 0.71 &  $\uparrow$ ObjectOnActivate,  $\downarrow$ Attacked,  $\downarrow$ PlayerShootingCorpse\\
RBQ & 21 & Is talkative & 0.20 & 3.13 &  $\downarrow$ Quest,  $\downarrow$ PlayerJumped, $\downarrow$ PlayerLootingCorpse \\
BPAQ & Physical & Physical Aggression & 0.19 & 27.9 &  $\uparrow$ PlayerDroppedItem,  $\uparrow$ PlayerShot,  $\uparrow$ PlayerSneaking\\
ACS & Temporal & Use of Temporal Dimension & 0.13 & 8.04 &  $\uparrow$ IteractionObject,  $\downarrow$ PlayerJumped,  $\downarrow$ Quest,\\
BPAQ & Anger & Anger & 0.11 & 63.70 &  $\uparrow$ PlayerShot,  $\uparrow$ PlayerShootingCorpse,  $\uparrow$ PlayerEquipped\\
RBQ & 62 & Speaks quickly & 0.10 & 1.15 &  $\uparrow$ Attack,  $\downarrow$ InteractionNPC,  $\uparrow$ PlayerDroppedItem \\
\hline
\end{tabular}
\end{table*}

\subsection{Data Collection Setup}
The experiment utilized the VPAL dataset, which had previously undergone a thorough review by the Institutional Review Board (IRB) and was approved for implementation. In total, 62 participants from the VPAL dataset were included in the study. Demographic information was extracted from the dataset, including their gender (32 male, 30 female), age, and occupation. The participants' ages ranged from 18 to 29 (M=20, SD=2.22) years, and they self-reported their game expertise as relatively experienced (M=3.89, SD=0.64) on a custom Likert scale from 1 to 5. The majority of participants were employed in various fields such as education, engineering, and healthcare, ensuring a diverse and representative sample for the study. It's important to note that the participants voluntarily participated in the experiment, and they were offered personality tests as compensation. Throughout the study, all participants were required to play the game in the laboratory for a duration of 60 minutes.
 
After this hour of playing, participants were asked to complete some of the most established questionnaires for the assessment of individual personality, such as the 300-item International Personality Item Pool Representation of the NEO-PI-R, 
the Ego-Resilience Scale (ER 89) \cite{b28f},
the Attributional Complexity Scale (ACS) \cite{b28d}, the Buss-Perry Aggression Questionnaire (BPAQ) \cite{b28e},
selected items of the Diagnostic and Statistical Manual of Mental Disorders (DSM-V) \cite{b28g}, brief measures of positive and negative affect through PANAS \cite{b28h}, and the Adult Repetitive Behaviours Questionnaire-2 (RBQ-2A) \cite{b28i}. The total duration of a session did not exceed two hours. Eventually, the complete data is publicly available through \textit{Game Data Science} \cite{GDS}.
\label{datacollection}

Before training a model, we first pre-processed, cleaned, and parsed the game logs to produce abstracted actions for each player (resulting in the activities of Table \ref{tab:in-game}). Secondly, a summary of all aggregated actions in the game was developed for each user, which was furthermore normalized based on the number of actions per minute. 

To uncover the individual reasoning behind the dialog choice making, we collected all meaningful in-game dialogs (i.e. those that offered the player a decision between affective expressions) and classified them into suitable categories after the taxonomy of conversation affect of Scott et al. (i.e.: \textit{Angry/Annoyed/Impatient, Supportive/Trust, Disbelief/Distraction, Apprehended/Afraid, Interest/Considering, Boredom/Frustration}) \cite{taxonomy}.

\section{Results}
To extract personality information from in-game behavior, random forest regression analysis was performed on 
the series of different factors from the formerly mentioned established personality measures (cf. Section \ref{datacollection}), following a common 80-20 training-testing split. Table \ref{tab:results} displays the most accurate predictions among these, ranked after their explained variance (R²) and the mean squared error (MSE) when compared to the testing set. For the sake of avoiding overinterpretation, we excluded all results that fell below the acceptable level of $R²=0.1$ in social sciences \cite{b13}.

Regarding the prediction from affective dialog choices, 
we followed the very same modeling approach, and present results in Table \ref{tab:resultsDialog}. 

\begin{table*}[h]
\centering
\caption{Most accurate personality variable predictions after random forest regression. We report the explained variance R², the mean squared error and the three most important features among dialog choices that predicted a factor. For the latter, we indicate the direction of correlation with the target variable.}
\label{tab:resultsDialog}
\begin{tabular}{|r|l|l|l|l|l|}
\hline
\multicolumn{1}{|c|}{\textbf{Scale}} &
\multicolumn{1}{|c|}{\textbf{Factor}} & \multicolumn{1}{c|}{\textbf{Description}} & \multicolumn{1}{c|}{\textbf{R²}} & \multicolumn{1}{c|}{\textbf{MSE}} &
\multicolumn{1}{c|}{\textbf{Highest Feature Importances}} 
\\
\hline

ACS & prefcomplex & Prefers complexity & 0.46 & 4.4 & $\downarrow$ Angry/Annoyed/Impatient, $\uparrow$ Supportive/Trust \\
RBQ & 54 & Speaks fluently and expresses ideas well & 0.29 &  1.11 & $\downarrow$ Disbelief/Distraction, $\downarrow$ Angry/Annoyed/Impatient, $\downarrow$ Apprehended/Afraid\\
RBQ & 34 & Tries to undermine, sabotage or obstruct & 0.25 & 0.47 & $\downarrow$ Apprehended/Afraid, $\uparrow$ Supportive/Trust, $\uparrow$ Interest/Considering\\
RBQ & 62 & Speaks quickly & 0.25 & 0.27 & $\uparrow$ Angry/Annoyed/Impatient, $\downarrow$ Boredom/Frustration, $\uparrow$ Disbelief/Distraction\\
RBQ & 12 & Physically animated & 0.24 & 0.42 & $\uparrow$ Interest/Considering, $\downarrow$ Angry/Annoyed/Impatient, $\downarrow$ Apprehended/Afraid\\
RBQ & 46 & Displays ambition & 0.24 & 1.23 & $\downarrow$ Boredom/Frustration, $\downarrow$ Supportive/Trust, $\downarrow$ Apprehended/Afraid \\
RBQ & 9 & Is reserved and unexpressive & 0.22 & 1.8 & $\uparrow$ Apprehended/Afraid, $\downarrow$ Supportive/Trust, $\uparrow$ Boredom/Frustration\\
RBQ & 28 & Exhibits condescending behavior & 0.22 & 1.39 & $\downarrow$ Apprehended/Afraid, $\uparrow$ Supportive/Trust, $\uparrow$ Interest/Considering\\
RBQ & 61 & Seems detached from the interaction & 0.22 & 1.14 & $\downarrow$ Sad/Apol./Embarrassed, $\uparrow$ Apprehended/Afraid, $\uparrow$ Interest/Considering\\
RBQ & 37 & Behaves in a Afraidful or timid manner & 0.21 & 1.55 & $\uparrow$ Sad/Apol./Embarrassed, $\uparrow$ Apprehended/Afraid, $\downarrow$ Angry/Annoyed/Impatient\\
RBQ & 63 & Acts playful & 0.21 & 0.8 & $\uparrow$ Supportive/Trust, $\uparrow$ Angry/Annoyed/Impatient, $\uparrow$ Interest/Considering\\
RBQ & 21 & Is talkative & 0.18 & 1.2 & $\uparrow$ Angry/Annoyed/Impatient, $\downarrow$ Sad/Apol./Embarrassed, $\downarrow$ Apprehended/Afraid\\
RBQ & 50 & Behaves in a cheerful manner & 0.17 & 1.08 & $\uparrow$ Supportive/Trust, $\downarrow$ Apprehended/Afraid, $\downarrow$ Angry/Annoyed/Impatient\\
FFM & A & Agreeableness & 0.16 & 140.5 & $\downarrow$ Boredom/Frustration, $\downarrow$ Apprehended/Afraid, $\uparrow$ Supportive/Trust \\
RBQ & 5 & Tries to control the interaction & 0.15 & 1.08 & $\downarrow$ Angry/Annoyed/Impatient, $\downarrow$ Boredom/Frustration, $\downarrow$ Apprehended/Afraid\\
RBQ & 10 & Laughs frequently & 0.11 & 1.3 & $\downarrow$ Supportive/Trust, $\uparrow$ Angry/Annoyed/Impatient, $\downarrow$ Apprehended/Afraid \\

\hline
\end{tabular}
\end{table*}

\section{Discussion}
The results from the aggregated data prediction indicated that the personality domain ``Openness'' (after FFM) had the highest R² value of 0.33 and a MSE of 214.80. 
Although explaining only a particular part of the variance, this prediction indicates that players who spend longer in-game time with sneaking (to pickpocket NPCs or creep up onto enemies for stealth attacks) show a reduced presence of Openness in their personality, which similarly holds for players that refrain from attacking NPCs first.
The tendency to collect (and use) a plenty of items and the aversion to drop them again plays well into the prediction of \textit{rigid perfection} from the DSM-V (R²=0.26).
The proneness to experiencing physical aggression (after BPAQ) correlates with players that interact in more risky game interaction, as reckless combat (getting shot) or sneaking onto NPCs. Even more drastically, BPAQ's Anger score could be predicted to a certain amount by players that have a higher tendency to shoot (defenseless) corpses, and getting shot themselves.

For the prediction from affective dialog choices, we could indicate that players who tend to show more supportive conversation with NPCs (and less anger or annoyance) rather prefer higher complexity (after ACS), which stands to reason with respect to complex gameplay, as this could be decremented with constantly aggressive expressions that either prevent more complex conversations with the NPC and/or even lead to crude combat instead of more complex quests.
Furthermore, players who showed less aggression, distraction or fear in their conversation are also connected to higher articulation skills (RBQ 54), and the expressed support and interest without apprehension could also be predictive for less benign intentions, such as undermining or sabotaging (RBQ 34).
Eventually, more certainly reasonable predictions could be found, such as RBQ 50 (behavior in cheerful manner) from supportive rather than fearful or angry dialog choices, Agreeableness (FFM) from support, afraidful and timid behavior (RBQ 37) from sad and afraid expressions, and players that are rather reserved and unexpressive (RBQ 9) also show that in higher apprehension and frustration, but less supportive statements.

% ...

% ...

\section{Limitations and Future Work}
While our study has shed light on the relationship between in-game actions in an RPG game and personality in gaming contexts, it is important to acknowledge its limitations and consider avenues for future research.
Even if a larger part of the predicted results stands to reason, some remain controversial, inconclusive or not directly backed up by theoretical background literature. For example, there is no intuitively reason why players that looted less items, equipped more items or dropped more items should show higher verbal aggression (after BPAQ) or why these should be predictive for that. While false negatives could be produced by algorithmic inaccuracy or a limitation of the data set, one has to acknowledge that, if deployed into actual usage, only statistically and theoretically justified connections should be considered (e.g. to tailor game content or classify players).

Specifically, studying individual differences is a challenging tasks due it's varied nature, and it requires consideration of many different factors. While this initial explorative study collected data from 64 participants, such datasets are relatively limited for training a model that addresses a highly noisy and multifarious field such as human personality, which resulted in low accuracy as compared to the prediction of more clear-cut variables. Collecting data from not only dozens, but rather millions of players constitutes our main endeavor for future work that we pursue in collaboration with an industrial game partner, which could challenge or even disprove the findings of this preliminary study. Future studies could also take advantage of supplementing other types of data, such as facial data and heart rate, as similar studies have indicated their impact on gaining a better understanding of human behavior, yet we refrained from this for the sake of scalability onto industrial usage. 
Another limitation of our work may be the lack of sufficient qualitative data, such as data-driven interviews conducted after the game. While we collected a significant amount of behavioral data, discussing in-game actions with the player could potentially improve the quality of the model and yield better results. While the RPG games utilized in this study were designed to cover different types of personalities and provide enough agency for players to experience and exhibit different behaviors in the game, they still suffer from a lack of specific tasks designed for individual differences and a validation method to gauge how players respond to and complete such tasks. For instance, tasks related to causality and reasoning within the game, as well as exploring the concept of intentionality between players, could provide more specific insights into individual differences and better validation of the results. 

By mere means of the breadth of the comparison and the big number of investigated personality variables, this study was prone to come up with certain significant correlations just because of the sheer number of predictions. Future work has to address and correct for this factor, yet in this stage we actively wanted to explore which scales are suitable to be predicted by in-game actions and dialog choices (e.g. RBQ turns out to be prominent). If deployed into actual (industrial) usage, such technology should considerably only be implemented when it is so mature that it serves the full range of personalities instead of only single factors.

Lastly, while the explained variance is already one indication of the predictive power of our approach, this investigation lacked a baseline to compare itself to. As we were more interested in the general possibility of using in-game actions and their impact on personality, we left the comparison to more sophisticated (yet explainable) machine learning models for future work.

\section{Conclusion}
In this paper, we utilized random forest regression and found that it yielded promising results for specific personality variables, thus demonstrating the feasibility of using such techniques in predicting in-game activities in theory.

By using these, and investigating even more advanced modeling techniques, we believe that we can create more accurate models for predicting in-game activities and potentially extend its use to other applications in the field of affective computing. Lastly, in order to further generalize our findings, we plan to extend our study to different genres and types of games as well as various cultural and demographic audiences, which could provide a more comprehensive understanding of the relationship between in-game actions and personality in gaming contexts, opening up new avenues for exploration in this exciting field of research.

\section*{Ethical Impact Statement}
First of all, as this approach is working on executed in-game data and conversational preferences, which can be highly sensitive and personal, the question of data ownership comes into play. Even if companies provide game environments and services, and therefore often have control over incoming and outgoing data, this data should ideally only be leveraged with the actively confirmed approval from the particular player (i.e. \textit{opt-in}). Ideally, echoing data transparency, players should have full insight and control over the history of their logs, so that unwanted entries could be permanently removed from storage and usage for the model. Moreover, even when being able to control their individual input, regular users can hardly estimate the impact of their data and how it could change in-game or higher-level decisions that certain use cases could determine - thus, in the spirit of explainability, users should be able to clearly follow the decisions of the model, its outcomes and implications on their experience with the product.
After all, modeling relationships between behavior, personality and affect and (algorithmically) deriving decisions from that should only be deployed for the benefit (e.g. improved experience) of the user, but bears the risk to be exploited to further facilitate dark patterns of (game) design, such as taking advantage of purchasing patterns or reinforcing addictive tendencies.
These risks excel in the case of erroneous decision making of the model, which could steer the individual's experience in the wrong direction or completely spoil it. Thus, if such a model is used for tailoring or adapting any element, it should only do if it can satisfy the prediction following a reasonable confidence.

\section*{Acknowledgements}
This work is funded by National Science Foundation (NSF), Division of Information and Intelligence Systems (IIS) under Grant Number 1355298 awarded to the project V-PAL: Virtual Personality Assessment
Laboratory. The authors would like to thank Alessandro Canossa, Jeremy B. Badler, Stefanie Tignor, C. Randall Colvin.

\vspace{12pt}
\color{red}

\end{document}